\pdfoutput=1

\documentclass[11pt]{article}

\usepackage[final]{coling}

\usepackage{times}
\usepackage{latexsym}
\usepackage{xcolor}
\definecolor{darkgreen}{rgb}{0.0, 0.5, 0.0}

\usepackage[T1]{fontenc}

\usepackage[utf8]{inputenc}

\usepackage{microtype}
\usepackage{amsmath}
\usepackage{multirow}
\usepackage{arydshln}
\usepackage{booktabs}
\usepackage{graphicx}
\usepackage{subfig}
\usepackage{xcolor}
\usepackage{hyperref}
\usepackage{listings}

\definecolor{codegreen}{rgb}{0,0.6,0}
\definecolor{codegray}{rgb}{0.5,0.5,0.5}
\definecolor{codepurple}{rgb}{0.58,0,0.82}
\definecolor{backcolour}{rgb}{0.95,0.95,0.92}

\lstdefinestyle{mystyle}{
    backgroundcolor=\color{backcolour},   
    commentstyle=\color{codegreen},
    keywordstyle=\color{magenta},
    numberstyle=\tiny\color{codegray},
    stringstyle=\color{codepurple},
    basicstyle=\ttfamily\footnotesize,
    breakatwhitespace=false,         
    breaklines=true,                 
    captionpos=b,                    
    keepspaces=true,                 
    numbers=left,                    
    numbersep=5pt,                  
    showspaces=false,                
    showstringspaces=false,
    showtabs=false,                  
    tabsize=2
}

\usepackage{inconsolata}

\usepackage{graphicx}

\title{CAST: Cross-modal Alignment Similarity Test for Vision Language Models}

\author{
  Gautier Dagan \\
  University of Edinburgh \\
  \texttt{gautier.dagan@ed.ac.uk} \\\And
  Olga Loginova \\
  University of Trento \\
  \texttt{olga.loginova@unitn.it} \\\And
  Anil Batra \\
  University of Edinburgh \\
  \texttt{a.k.batra@sms.ed.ac.uk} \\
  }

\begin{document}
\maketitle
\begin{abstract}

Vision Language Models (VLMs) are typically evaluated with Visual Question Answering (VQA) tasks which assess a model's understanding of scenes. 
Good VQA performance is taken as evidence that the model will perform well on a broader range of tasks that require both visual and language inputs. 
However, scene-aware VQA does not fully capture input biases or assess hallucinations caused by a misalignment between modalities.
To address this, we propose a Cross-modal Alignment Similarity Test (CAST) to probe VLMs for \textit{self-consistency} across modalities. 
This test involves asking the models to identify similarities between two scenes through text-only, image-only, or both and then assess the truthfulness of the similarities they generate. 
Since there is no ground-truth to compare against, this evaluation does not focus on objective accuracy but rather on whether VLMs are internally consistent in their outputs. 
We argue that while not all self-consistent models are capable or accurate, all capable VLMs must be self-consistent.

\end{abstract}
\section{Introduction}
\label{sec:intro}

Vision Language Models (VLMs) integrate vision and language modalities to learn image-text correspondences from large-scale image-text pairs \cite{Zhang2023VisionLanguageMF, Radford2021LearningTV, Kwon2022MaskedVA}. 
Given image-text pairs, VLMs combine a text encoder and an image encoder to extract image and text features and then learn to align vision and language through generative objectives, such as Visual Question Answering (VQA). 
As a result, VLMs pose a unique challenge in ensuring consistent outputs across different input types – be it text, images, or a combination of both.

Consistency in AI models is essential for their reliability and trustworthiness \cite{Ji2023AIAA}. 
Self-consistency refers to a model's ability to produce stable, coherent outputs across similar inputs and conditions \cite{elazar-etal-2021-measuring}. 
If a VLM exhibits inconsistent behavior when given the same input across different modalities, it could raise concerns about its robustness and internal reasoning. 
So while models might perform well on major VQA benchmarks, such as MMMU \cite{Yue2023MMMUAM} or MME \cite{Fu2023MMEAC}, we argue that they must also be evaluated for self-consistency.

\begin{figure}[t!]
    \centering
    \includegraphics[width=0.48\textwidth]{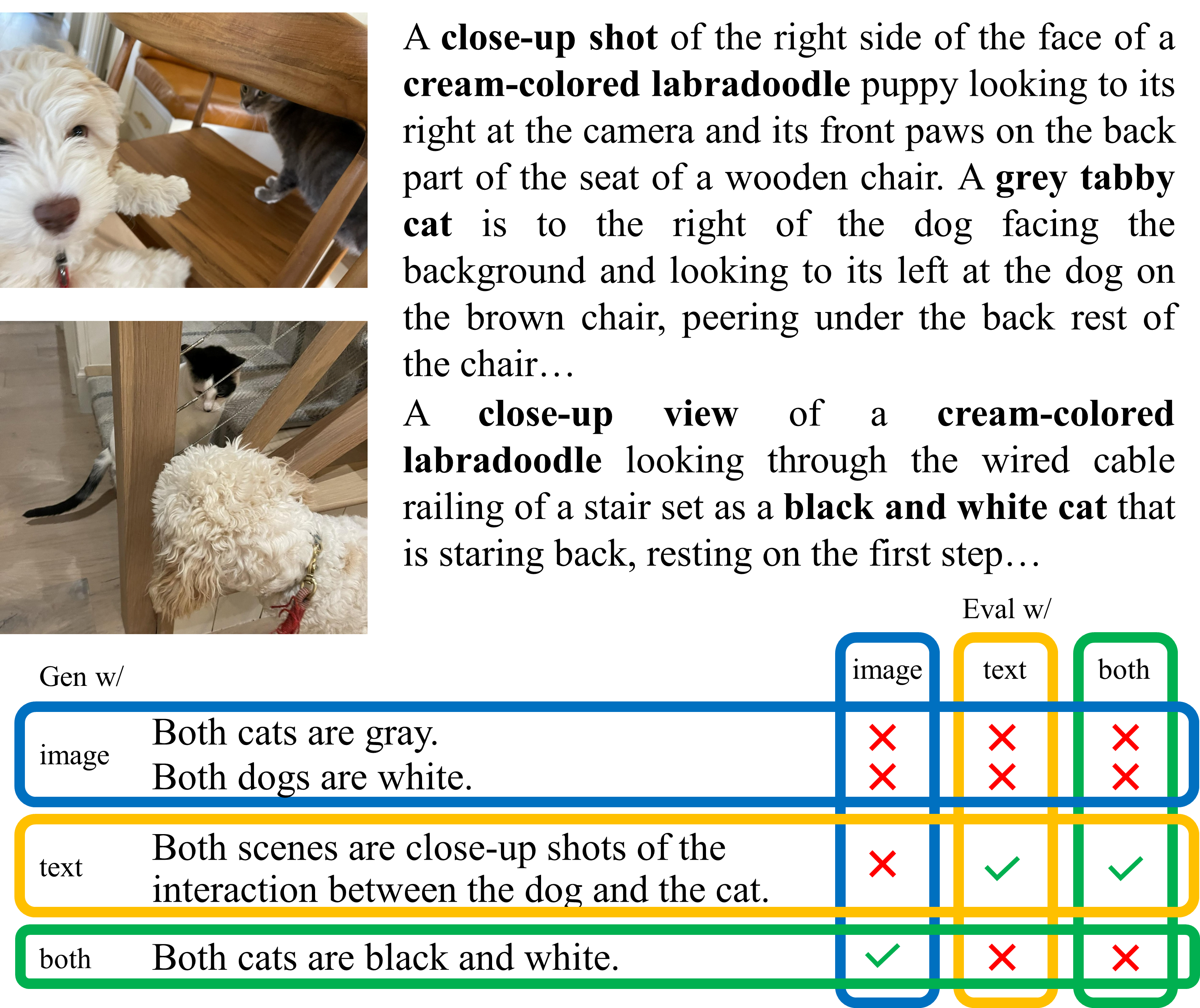}
    \caption{Example of paired scenes and statements from the CAST dataset. Horizontal blocks show generated statements, while vertical blocks are evaluations for each modality: \textcolor[HTML]{2f6eba}{image-only}, \textcolor[HTML]{f09737}{text-only}, and \textcolor[HTML]{4ead5b}{image+text}. \textcolor{red}{Red} crosses indicate where each model disagrees with its own generation during the evaluation step. Similarity topics  are highlighted \textbf{in bold}. 
    Note that VLMs may produce hallucinations, as the CAST method checks for consistency rather than correctness.}
    \label{fig:example}
\end{figure}

We propose to evaluate self-consistency through the absence of contradictions between a model's generated output and the evaluation of this output by different modalities. 
To this end, we introduce the two-step Cross-modal Alignment Similarity Test (CAST).
\footnote{We publicly release all our code and dataset \href{https://github.com/gautierdag/cast}{here}.}
We apply CAST to different VLMs and find that despite strong performance on many other downstream tasks, the majority of VLMs exhibit a lack of internal self-consistency and modality alignment (see examples in Figure~\ref{fig:example}).
CAST provides a more nuanced understanding of VLMs' reasoning capabilities and potential biases, which is critical for real-world applications.

\begin{figure*}[ht]
    \centering
    \includegraphics[width=\linewidth]{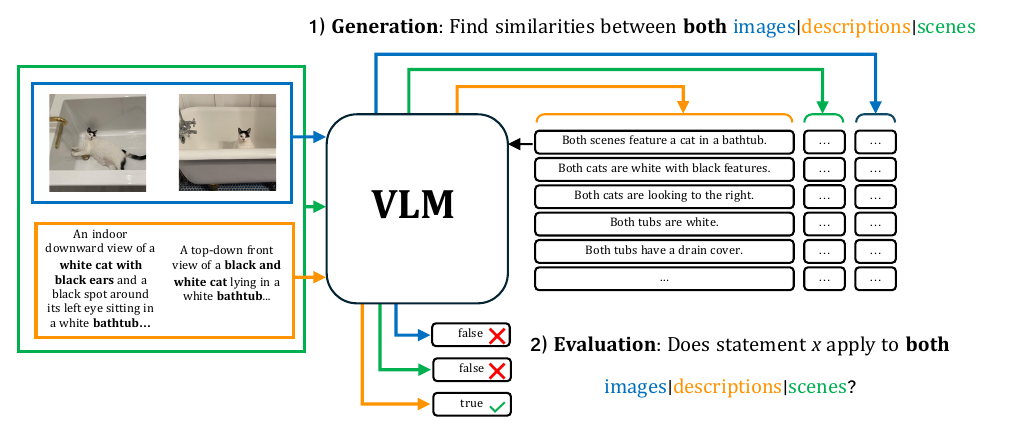}
    \caption{CAST is two-fold. 
    In the first step, we ask the model to \textit{generate} a set of similarity statements conditioned on different modality input types (image-only, text-only, both). 
    In the second step, the model \textit{validates} the truthfulness of the generated statements with respect to each modality. This allows us to measure whether the VLM is self-consistent within a modality and across different modalities. 
    }
    \label{fig:method}
\end{figure*}

\section{Related Works}
\label{sec:rw}

Traditionally, self-consistency has been tested through meaning-preserving alternations to model's inputs, such as adding illogical statements, filler tokens, or paraphrasing \cite{elazar-etal-2021-measuring, Parcalabescu2024DoV, yue2024sc}. 
Logical consistency ensures that the model's outputs remain coherent and non-contradictory throughout multiple iterations \cite{Yang2024DecomposeAC,Zhang2024UnveilingTT}.
We instead design CAST to evaluate the \textbf{cross-modal} consistency of VLMs through a comparison task.

Several work have proposed image comparison benchmarks for VLMs \cite{Fu2024BLINKML, Zhao2024BenchmarkingMU, Dunlap2023DescribingDI}. They focus on contrastive pairs where similar images differ in key features. For example, VisDiffBench \cite{Dunlap2023DescribingDI} uses humanannotated differences between two sets of images. 
In contrast, we focus on similarities to capture the semantic overlap between example pairs.

Self-consistency in VLMs and LLMs is closely tied to uncertainty in predictions, resulting in noisier outputs \cite{chen2024inside}. Consequently, it is used as a metric to detect hallucinations caused by misalignment \cite{manakul-etal-2023-selfcheckgpt,mundler2024selfcontradictory,Li2024ReferencefreeHD}. 
CAST can also reveal logical hallucinations where a model's uncertainty causes it to be inconsistent.

\section{Method}
\label{sec:method}

We propose CAST (shown in Figure~\ref{fig:method}) as a fully-automated two-step approach to evaluate multi-modal self-consistency in VLMs.

\subsection{Generating Similarities}

CAST leverages similarities between two scenes to assess a model's ability to evaluate its own outputs. 
In our case, a scene is an image paired with its high-quality description (see Section~\ref{sec:dataset-main}).
By focusing on shared features, the model is less likely to rely on surface-level distinctions or superficial strategies. 
For instance, if tasked with finding differences between two images, the model might only attend to one image or highlight minor details like color changes. 
Emphasizing similarities encourages a deeper evaluation of each input.

The first step of CAST is to prompt the VLM to generate a number of statements about the similarities between two input scenes $S_{A}$ and $S_{B}$.
Since we generate a list of similarities using the VLM, each subsequent similarity statement is conditioned on all previously generated ones. 
We can view the generation of a given similarity as following:
\begin{align}
sim_{0} &= VLM(S_{A}, S_{B}, P^{gen}) \\    
sim_{i} &= VLM(sim_{i-1},...,sim_{0}; S_{A}, S_{B}, P^{gen}),
\end{align} where $P^{gen}$ is the instructions.
Similarity statements are generated for different modalities: scenes can be represented as two images ($S^{img}$), two text descriptions ($S^{txt}$), or two images combined with the corresponding descriptions ($S^{img+txt}$).
We obtain the similarity statements conditioned on a pair of scenes for each stream.
We restrict the input pairs to the same modality, and generate all statements using greedy sampling ($t=0$).

\subsection{Evaluating Similarities}

The second step of our approach is to evaluate each similarity statement and test whether a model remains consistent under different modalities.
Since we focus on self-consistency, we use the same model for both generation and evaluation.
The evaluation step can be represented as following:
\begin{align}
s &= VLM(S_{A}, S_{B}, P^{eval}), 
\end{align} where $s$ is $1$ if the model confirms that the statement is true and $0$ otherwise.
We filter out the generations that cannot be parsed (see Appendix~\ref{sec:prompts} for details).
To mitigate bias towards a certain prompt or phrasing \cite{prompt-order, sclar2024quantifying}, we use three different evaluation prompts.
Thus, apart from the conventional Yes/No questions, we ask the model whether the statement applies to one or both scenes and whether the statement is true or false.
To quantify self-consistency, we report the average $s$ over all the evaluated pairs and prompts for each modality permutation (both generated and evaluated with).

\section{Experiments and Results}
\label{sec:exp}

\subsection{Dataset}
\label{sec:dataset-main}

Since CAST relies on asking VLMs to find similarities between two scenes, we need a multi-modal set of pairs of aligned images/descriptions that contain similarities. 
To construct our evaluation dataset, we sub-sample example pairs from the DOCCI Dataset \cite{onoe2024docci}.
The dataset contains 15k images paired with human-annotated descriptions of 136 words on average. 
The images focus on spatial relations and world knowledge. 
Unlike popular captioning datasets, each description is comprehensively annotated to capture the differences between similar images.

We randomly sample 100 pairs of images from the DOCCI train dataset of 10k images.
We threshold the CLIP \cite{clip} cosine-similarity and filter out the pairs of the $<0.75$ CLIP score (since the images might not have enough in common), or $\geq 0.95$ CLIP score (to exclude near identical ones and duplicates).
We also filter out images with captions of less than 500 characters to include only those that contain ample descriptive information about the scene.

\subsection{Models}

We test the following open-source and closed-source VLMs for self-consistency, each with distinct vision encoders, language models, and training dataset:
\begin{itemize}
    \setlength\itemsep{-0.5em}
    \item \textbf{Bunny 1.1}~\cite{he2024bunny}
    \item \textbf{LLaVA}~\cite{liu2023improved} in three configurations: LLaVA 1.5 (Vicuna), LLaVA 1.6 (Llama), and LLaVA 1.6 (Mistral). Additionally, we evaluate LLaVA 1.5 RLAIF \cite{yu2024rlaif}, a version of LLaVA 1.5 aligned through AI feedback.
    \item \textbf{InternVL2}~\cite{chen2023internvl}
    \item \textbf{MiniCPM V2}~\cite{yao2024minicpmv}
    \item \textbf{Phi 3.5 Vision}~\cite{abdin2024phi}
    \item \textbf{GPT4o-mini}
\end{itemize}
See Appendix~\ref{sec:additional-model-details} for more information on each model.

\subsection{Results}

\begin{table}[t]
\centering
\tiny
    \begin{tabular}{ccccc}
        \toprule
         \textbf{Model} & \textbf{Gen w/} & \textbf{Eval w/ text} & \textbf{Eval w/ image} & \textbf{Eval w/ both} \\
         \midrule
         \multirow{3}{*}{Bunny} & text & \textbf{0.93} & 0.76 & 0.96 \\
                                & image & 0.71 & \textbf{0.80} & 0.85 \\
                                & both & 0.91 & 0.81 & \textbf{0.96} \\
        \cline{1-5}
        \multirow{3}{*}{GPT4o-Mini} & text & \textbf{0.94} & 0.73 & 0.94 \\
                                    & image & 0.79 & \textbf{0.90} & 0.87 \\
                                    & both & 0.91 & 0.76 & \textbf{0.91} \\
        \cline{1-5}
        \multirow{3}{*}{InternVL2} & text & \textbf{0.67} & 0.66 & 0.72 \\
                                    & image & 0.57 & \textbf{0.78} & 0.75 \\
                                    & both & 0.68 & 0.73 & \textbf{0.77} \\
        \cline{1-5}
        \multirow{3}{*}{MiniCPM V2} & text & \textbf{0.92} & 0.90 & 0.93 \\
                                    & image & 0.50 & \textbf{0.91} & 0.73 \\
                                    & both & 0.84 & 0.89 & \textbf{0.91} \\
        \cline{1-5}
        \multirow{3}{*}{Phi-3.5-V} & text & \textbf{0.61} & 0.60 & 0.63 \\
                                    & image & 0.50 & \textbf{0.72} & 0.61 \\
                                    & both & 0.60 & 0.63 & \textbf{0.64} \\
        \cline{1-5}
        \multirow{3}{*}{LLaVA 1.5 (Vicuna)} & text & \textbf{0.91} & 0.87 & 0.82 \\
                                    & image & 0.68 & \textbf{0.91} & 0.69 \\
                                    & both & 0.86 & 0.87 & \textbf{0.78} \\
        \cline{1-5}
        \multirow{3}{*}{LLaVA 1.6 (Llama)} & text & \textbf{0.73} & 0.57 & 0.74 \\
                                & image & 0.56 & \textbf{0.69} & 0.64 \\
                                & both & 0.68 & 0.62 & \textbf{0.73} \\
        \cline{1-5}
        \multirow{3}{*}{LLaVA 1.6 (Mistral)} & text & \textbf{0.81} & 0.85 & 0.87 \\
                                    & image & 0.52 & \textbf{0.86} & 0.72 \\
                                    & both & 0.74 & 0.86 & \textbf{0.84} \\
        \cline{1-5}
         \multirow{3}{*}{LLaVA 1.5 RLAIF} & text & \textbf{0.57} & 0.66 & 0.49 \\
                                & image & 0.58 & \textbf{0.93} & 0.70 \\
                                & both & 0.55 & 0.76 & \textbf{0.58} \\
        \bottomrule
    \end{tabular}
    \caption{CAST self-consistency scores (Top-3) averaged over the first three statements generated for each modality configuration.
    Bold cells show the performance when the evaluation is in the same modality as the generation.}
\label{tab:results}
\end{table}

Table \ref{tab:results} shows the CAST results for similarity statements generated and evaluated across different modalities.
We average the CAST score over the first the first three statements generated (Top-3).
The results indicate that models perform best when statements are generated and evaluated within the same modality.
There is a noticeable drop in consistency during cross-modal evaluations, where statements generated from images are evaluated using text descriptions and vice versa. 
With the exception of GPT4o-Mini, the combination of image-generated and text-evaluated statements leads to the worst consistency.
This is somewhat expected as the similarity statement generated by the model might have been about something not mentioned in the text description (see Appendix~\ref{sec:add-info}).

Qualitatively, we find that most inconsistencies arise during generation, where models often produce incorrect statements, particularly about object attributes or relationships. 
Notably, the image modality shows the highest hallucination rates, with models emphasizing prominent features without verifying their relevance to both scenes. 
This suggests that while object recognition is strong in state-of-the-art VLMs, accurately describing attributes and relations remains a challenge.

MiniCPM exhibits high consistency when evaluating with images. 
To test whether this is due to its RLAIF \cite{yu2024rlaif} fine-tuning stage, we evaluate a version of LLaVA-1.5 specially trained with RLAIF.
Overall, we find LLaVA-1.5 RLAIF to be significantly less consistent than its base-model LLaVA-1.5.
We therefore fail to conclude that RLAIF has a positive impact on consistency.

\begin{figure}
    \centering
    \includegraphics[width=0.48\textwidth]{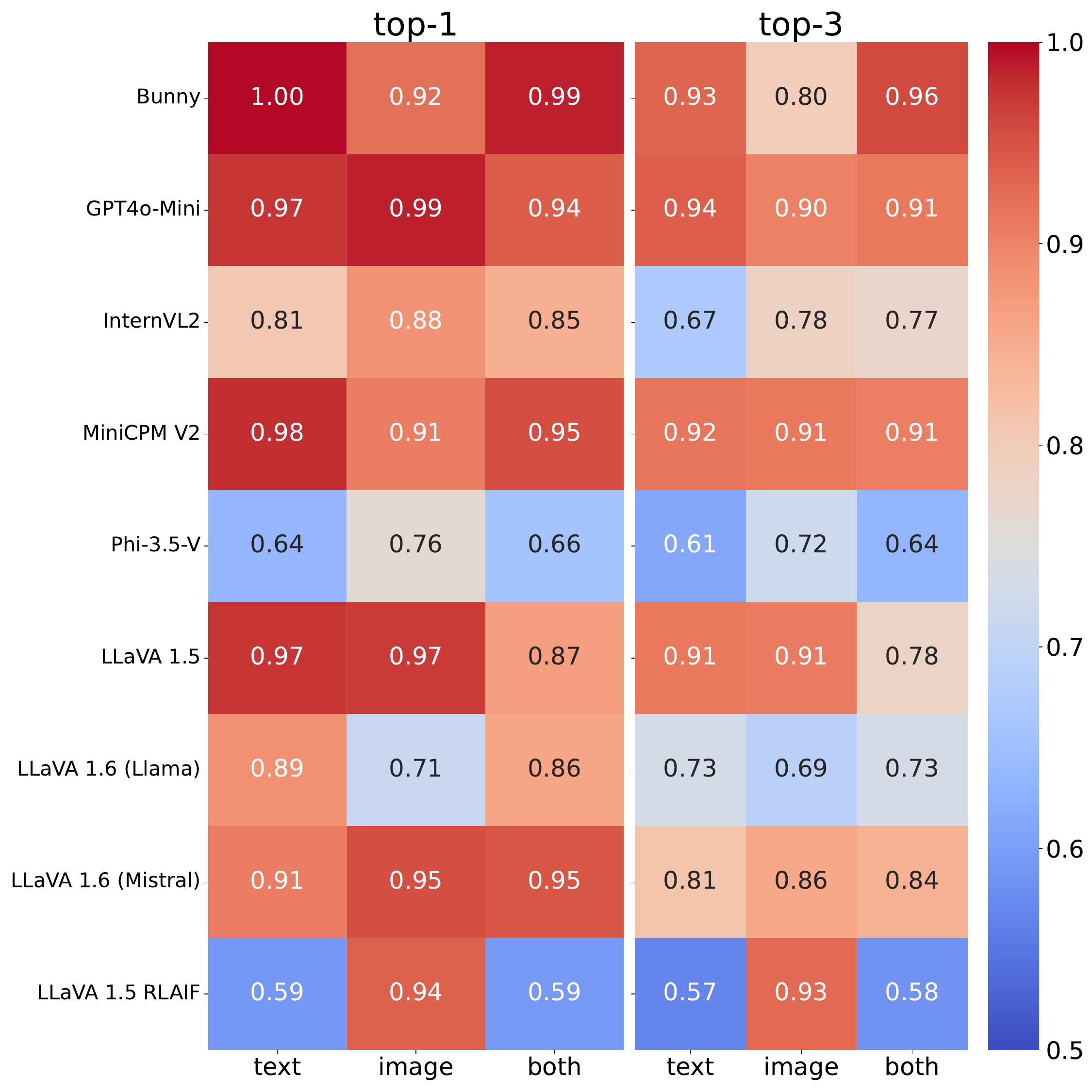}
    \caption{Average CAST self-consistency when multiple statements are generated and evaluated within the same modality. \textbf{Left}: Top-1 considers only the first statement generated. \textbf{Right}: Top-3 considers the first three statements generated, these are equivalent to the bolded results from Table~\ref{tab:results}. }
    \label{fig:model_consistency}
\end{figure}

Figure~\ref{fig:model_consistency} shows CAST scores for Top-1 and Top-3 generated statements. There is a slight decrease in CAST scores from Top-1 to Top-3, indicating that the quality of similarity statements typically declines with additional generations, as models become less reliable over longer generations.

We find GPT4o-Mini and MiniCPM to be the most consistent models overall.
Both exhibit minimal drop with longer generations ($9\%$ for GPT4o-Mini and $6\%$ for MiniCPM). 
In contrast, InternV2 and the LLaVA models experience a significant drop in consistency with additional generations.
Overall, our single-modality CAST results highlight that VLMs fail to provide coherent and stable outputs as generations get longer.

Lastly, we can use CAST to evaluate how different modalities impact different VLMs. 
For instance, GPT4o-Mini and Bunny show a drop in image self-consistency when generation length increases, unlike MiniCPM and LLaVA-RLAIF which maintain more stability with generation length. 
Other models such as InterVL2 are more sensitive to the text modality.

\section{Conclusion}
\label{sec:concl}

We introduce CAST to evaluate the multi-modal self-consistency of VLMs by testing whether a model applies consistent reasoning across text-only, image-only, or combined inputs. 
CAST uncovers cross-modal inconsistencies and goes beyond traditional accuracy metrics to assess the stability of a model's logic across different modalities.

Our findings show that open-source VLMs still struggle with self-consistency across different modalities.
CAST not only assesses self-consistency but also identifies modalities where the model may lack understanding. 
While we do not evaluate model correctness, we emphasize that self-consistency is a crucial, yet often overlooked, metric in multi-modal AI systems, essential for improving robustness. 

Furthermore, given the method's universality, CAST's framework can be adapted to any domain or language dataset, provided there are sufficiently similar images and highly detailed descriptions.

\section{Limitations}
\label{sec:lim}

The main limitation is that our test does not guarantee the capability of a model. 
We make no claims about the correctness of the model, but focus solely on whether a model is self-consistent. 
This means a model that always predicts the similarity statement to match the scenes, regardless of the statement, would always be deemed consistent even though it would also likely be wrong. 
Our approach therefore needs to be taken in conjunction with the traditional evaluation methods. It is most useful for models trained and evaluated using standard correctness metrics. 

There are also limitations with our VLM evaluations that follow directly from the brittle nature of these models. 
While we evaluated the generated statements using multiple prompts, we sample from each model using greedy sampling and therefore it is possible that some of our results are biased towards certain models.
However, CAST could easily be expanded to include responses from different sampling mechanisms (temperature > 0) at the cost of increased computation.

\section{Ethical Considerations}

Our research relies on open-source and closed-source VLMs generating and evaluating text and image inputs and therefore carries the typical risks associated with open-ended text generation.
The DOCCI dataset, which we sub-sample from, is licensed under the CC BY 4.0 license\footnote{\href{https://creativecommons.org/licenses/by/4.0/}{https://creativecommons.org/licenses/by/4.0/}}.
Overall, we hope that CAST leads to improvements in the trustworthiness and robustness of VLMs.

\section*{Acknowledgments}

 The authors extend their gratitude to Amazon's Development Centre Scotland (ADCS) for the challenge and AWS access to work with VLMs models of the Claude family. In particular, we wish to thank Christos Christodoulopoulos for his helpful feedback throughtout this project. 

This work was supported in part by the UKRI Centre for Doctoral Training in Natural Language Processing, funded by the UKRI (grant EP/S022481/1) at the University of Edinburgh, School of Informatics and School of Philosophy, Psychology \& Language Sciences and by the UKRI-funded TAS Governance Node (grant number EP/V026607/1).

\bibliography{custom}

\clearpage
\appendix
\section{Prompts}
\label{sec:prompts}

\subsection{Generation}

To generate a number of similarity statements, we use the prompt shown in Figure~\ref{fig:gen-prompt}. 
We slightly modify the prompt to fit each modality input.

\begin{figure}[h]
\texttt{Given two \textcolor[HTML]{f09737}{scenes} | \textcolor[HTML]{2f6eba}{side-by-side images} | \textcolor[HTML]{4ead5b}{scenes and their corresponding images}, find up to five similarities between each \textcolor[HTML]{f09737}{scene}|\textcolor[HTML]{2f6eba}{image}|\textcolor[HTML]{4ead5b}{scene}. Output each similarity in a numbered list.}
\caption{Generation Prompt: For each model and each of the three modalities, we generate a list of similarity statements using the above prompt.}
\label{fig:gen-prompt}
\end{figure}

\subsection{Evaluation}

To reduce variance in our results and potential biases that might exist towards certain prompt phrasing \cite{prompt-order, sclar2024quantifying}, we opt to use three different evaluation prompts shown in Figure~\ref{fig:eval-prompt}.

\begin{figure}[h]
\begin{enumerate}
    \item \texttt{Given two \textcolor[HTML]{f09737}{scenes}|\textcolor[HTML]{2f6eba}{side-by-side images}|\textcolor[HTML]{4ead5b}{scenes and their corresponding images}, does the following statement apply to only one of the \textcolor[HTML]{f09737}{scenes} | \textcolor[HTML]{2f6eba}{images} | \textcolor[HTML]{4ead5b}{scenes}? Answer with `one' or `both'.}

    \item \texttt{Given two \textcolor[HTML]{f09737}{scenes}|\textcolor[HTML]{2f6eba}{side-by-side images}|\textcolor[HTML]{4ead5b}{scenes and their corresponding images}, is the following statement true for both of the \textcolor[HTML]{f09737}{scenes}|\textcolor[HTML]{2f6eba}{images}|\textcolor[HTML]{4ead5b}{scenes}? Answer with `true' or `false' if the statement is untrue or only true for one of the \textcolor[HTML]{f09737}{scenes}|\textcolor[HTML]{2f6eba}{images}|\textcolor[HTML]{4ead5b}{scenes}.}

    \item \texttt{Given two \textcolor[HTML]{f09737}{descriptions}|\textcolor[HTML]{2f6eba}{side-by-side images}|\textcolor[HTML]{4ead5b}{descriptions and their corresponding images}, does the following statement describe both of the \textcolor[HTML]{f09737}{descriptions}|\textcolor[HTML]{2f6eba}{images}|\textcolor[HTML]{4ead5b}{descriptions}? Answer with `yes' or `no' if the statement is not applicable to one of the \textcolor[HTML]{f09737}{descriptions}|\textcolor[HTML]{2f6eba}{images}|\textcolor[HTML]{4ead5b}{descriptions}.}
\end{enumerate}
\caption{Evaluation Prompts: For each model and each of the three modalities, we generate validate a similarity statement from the generation step. We use three different evaluation prompts to reduce potential bias of models towards a particular prompt format.}
\label{fig:eval-prompt}
\end{figure}

\subsection{Parsing the Evaluation Output}

To parse the resulting evaluation from the model we use a simple post-processing step:

\begin{lstlisting}[language=Python]
def parse_validator(x):
    x = x.strip("*").lower().split("\n")[0]
    if x.startswith(positive):
        return 1
    elif x.startswith(negative):
        return 0
    else:
        return None
\end{lstlisting}

Note that we ignore generations that we cannot parse from the evaluation score.
This is typically rare for most models and prompts.

\section{Additional model details}
\label{sec:additional-model-details}

\begin{table*}[]
    \centering
    \scriptsize
    \scalebox{0.9}{
    \begin{tabular}{|c|c|c|c|}
        \hline 
         Model & Vision Encoder & LLM & Additional Design Choices \\
         \hline
         Bunny 1.1 \cite{he2024bunny} & SigLip-400M \cite{siglip} & Llama-3-8B Ins \cite{dubey2024llama} & -  \\
         MiniCPM V 2.5 \cite{yao2024minicpmv}  & SigLip-400M \cite{siglip} & Llama-3-8B Ins \cite{dubey2024llama} & Adaptive Visual Encoding and RLAIF-V \\
         InternVL2 \cite{chen2023internvl} & InternViT & InternLM 2.5 7B & - \\
         Phi 3.5 Vision \cite{abdin2024phi} & CLIP ViT \cite{clip} & phi-3-mini-128K-instruct LM & - \\
         LLaVa-Next 1.5 \cite{liu2023llava} & CLIP-ViT \cite{clip} & Vicuna-7B  & - \\
         LLaVa-Next 1.6 \cite{liu2023llava} & CLIP-ViT \cite{clip} & Mistral-7B \cite{jiang2023mistral} & - \\
         LLaVa-Next 1.6 \cite{liu2023llava} & CLIP-ViT \cite{clip} & Llama-3-8B Ins \cite{dubey2024llama} & Image Slicing \\
         LLaVa-Next 1.5 RLAIF \cite{yu2024rlaif} & CLIP-ViT \cite{clip} & Vicuna-7B &  Visual RLAIF Alignment\\
         \hline
    \end{tabular}}
    \caption{Open-source VLMs tested along with a description of which Vision Encoder and LLM each model uses.}
    \label{tab:model-compare}
\end{table*}

\section{Additional Results}
\subsection{Results for each Prompt types}
\begin{table*}[t]
\centering
\tiny
    \begin{tabular}{cccccccccccccc}
        \toprule
         \multirow{2}{*}{\textbf{Model}} & \multirow{2}{*}{\textbf{Gen w/}} & \multicolumn{4}{c}{\textbf{Eval w/ text}} & \multicolumn{4}{c}{\textbf{Eval w/ image}} & \multicolumn{4}{c}{\textbf{Eval w/ both}} \\
         & & yes/no & both/one & true/false & Avg. & yes/no & both/one & true/false & Avg. & yes/no & both/one & true/false & Avg. \\
         \midrule
         \multirow{3}{*}{Bunny} & text & 0.97 & 0.99 & 0.84 & 0.93 & 0.73 & 0.88 & 0.66 & 0.76 & 0.99 & 0.96 & 0.93 & 0.96 \\
 & image & 0.72 & 0.86 & 0.55 & 0.71 & 0.75 & 0.90 & 0.74 & 0.80 & 0.85 & 0.89 & 0.80 & 0.85 \\
 & both & 0.94 & 0.98 & 0.80 & 0.91 & 0.77 & 0.92 & 0.74 & 0.81 & 0.98 & 0.96 & 0.94 & 0.96 \\
         \cline{1-14}
        \multirow{3}{*}{GPT4o-M} & text & 0.91 & 0.98 & 0.92 & 0.94 & 0.70 & 0.88 & 0.62 & 0.73 & 0.96 & 0.95 & 0.93 & 0.94 \\
 & image & 0.75 & 0.91 & 0.71 & 0.79 & 0.91 & 0.95 & 0.85 & 0.90 & 0.86 & 0.94 & 0.83 & 0.87 \\
 & both & 0.90 & 0.97 & 0.86 & 0.91 & 0.73 & 0.87 & 0.67 & 0.76 & 0.93 & 0.94 & 0.87 & 0.91 \\
        \cline{1-14}
        \multirow{3}{*}{InternVL2} & text & 0.80 & 0.36 & 0.86 & 0.67 & 0.86 & 0.17 & 0.94 & 0.66 & 0.90 & 0.32 & 0.95 & 0.72 \\
 & image & 0.65 & 0.34 & 0.71 & 0.57 & 0.98 & 0.39 & 0.99 & 0.78 & 0.88 & 0.42 & 0.95 & 0.75 \\
 & both & 0.85 & 0.35 & 0.85 & 0.68 & 0.96 & 0.25 & 0.99 & 0.73 & 0.94 & 0.40 & 0.98 & 0.77 \\
         \cline{1-14}
         \multirow{3}{*}{MiniCPM} & text & 0.96 & 0.79 & 0.99 & 0.92 & 0.96 & 0.78 & 0.97 & 0.90 & 0.98 & 0.83 & 0.99 & 0.93 \\
 & image & 0.48 & 0.40 & 0.61 & 0.50 & 0.96 & 0.82 & 0.96 & 0.91 & 0.73 & 0.61 & 0.84 & 0.73 \\
 & both & 0.89 & 0.69 & 0.95 & 0.84 & 0.93 & 0.78 & 0.95 & 0.89 & 0.96 & 0.78 & 0.98 & 0.91 \\
        \cline{1-14}
        \multirow{3}{*}{Phi-V} & text & 0.81 & 0.09 & 0.94 & 0.61 & 0.82 & 0.12 & 0.85 & 0.60 & 0.89 & 0.12 & 0.87 & 0.63 \\
         & image & 0.58 & 0.20 & 0.73 & 0.50 & 0.91 & 0.32 & 0.94 & 0.72 & 0.82 & 0.24 & 0.75 & 0.61 \\
         & both & 0.74 & 0.18 & 0.87 & 0.60 & 0.85 & 0.18 & 0.87 & 0.63 & 0.89 & 0.16 & 0.86 & 0.64 \\
        \cline{1-14}
        \multirow{3}{*}{LLaVA1.5} & text & 0.76 & 0.99 & 0.98 & 0.91 & 0.66 & 1.00 & 0.95 & 0.87 & 0.50 & 0.98 & 0.97 & 0.82 \\
 & image & 0.40 & 0.89 & 0.74 & 0.68 & 0.75 & 1.00 & 0.98 & 0.91 & 0.33 & 0.92 & 0.83 & 0.69 \\
 & both & 0.64 & 0.99 & 0.94 & 0.86 & 0.65 & 1.00 & 0.95 & 0.87 & 0.42 & 0.98 & 0.94 & 0.78 \\
         \cline{1-14}
         \multirow{3}{*}{LLaVA-1.6 (Llama)} & text & 0.69 & 0.99 & 0.51 & 0.73 & 0.48 & 1.00 & 0.24 & 0.57 & 0.71 & 0.99 & 0.52 & 0.74 \\
 & image & 0.42 & 0.93 & 0.33 & 0.56 & 0.72 & 0.99 & 0.36 & 0.69 & 0.54 & 0.96 & 0.40 & 0.64 \\
 & both & 0.65 & 0.98 & 0.43 & 0.68 & 0.58 & 0.99 & 0.29 & 0.62 & 0.71 & 0.99 & 0.50 & 0.73 \\
        \cline{1-14}
        \multirow{3}{*}{LLaVA1.6 (Mistral)} & text & 0.89 & 0.65 & 0.89 & 0.81 & 0.88 & 0.78 & 0.90 & 0.85 & 0.95 & 0.71 & 0.95 & 0.87 \\
 & image & 0.59 & 0.37 & 0.59 & 0.52 & 0.89 & 0.74 & 0.93 & 0.86 & 0.79 & 0.55 & 0.81 & 0.72 \\
 & both & 0.82 & 0.57 & 0.84 & 0.74 & 0.90 & 0.76 & 0.93 & 0.86 & 0.93 & 0.67 & 0.93 & 0.84 \\
        \cline{1-14}
        \multirow{3}{*}{LLaVA1.5 RLAIF} & text & 0.19 & 0.86 & 0.66 & 0.57 & 0.37 & 1.00 & 0.62 & 0.66 & 0.22 & 0.89 & 0.35 & 0.49 \\
         & image & 0.28 & 0.92 & 0.53 & 0.58 & 0.84 & 1.00 & 0.95 & 0.93 & 0.49 & 0.95 & 0.66 & 0.70 \\
         & both & 0.22 & 0.89 & 0.55 & 0.55 & 0.54 & 1.00 & 0.74 & 0.76 & 0.33 & 0.92 & 0.50 & 0.58 \\
        \bottomrule
    \end{tabular}
    \caption{CAST self-consistency scores for the first three statements generated for each modality configuration.}
\label{tab:results}
\end{table*}

\subsection{Results for position of generated statement}

\section{Information Flow in Image Similarity}
\label{sec:add-info}

Since the human annotators of DOCCI \cite{onoe2024docci} are given the image from which to write the description, from an information content, we can view the textual description of an image as a subset of the overall information content contained within the image.

If denote information content as entropy $H$, then:
$$
H(S^{img}) \geq H(S^{txt})
$$

And since the text description should not be introducing new information, the union of both the Image and Description should be equal in entropy to that of the image:

$$
H(S^{img}) = H(S^{img+txt})
$$

Unfortunately, it is the case that text can introduces new information through subjective interpretation, and the obvious fact that a photograph can very rarely be fully described in language.
However it might still be useful to model the annotation of images as conditional on the images and not independent. 
This might lead to further inter-modal consistency analysis which we leave open as direction for future work.

\section{Dataset Sub-sampling}
\label{sec:dataset}

As previously mentioned, we sample image pairs from the DOCCI \cite{onoe2024docci} train dataset (10k images), and reject pairs that do not exhibit a certain threshold of CLIP similarity.
In particular, we use cosine-similarity between images to filter pairs which are either not similar enough ($<0.75$ CLIP score), or which contain near identical images or duplicates ($\geq 0.95$ CLIP score). 
We decided on these boundaries through qualitative analysis of the DOCCI samples.
Additionally, we filter pairs by description length to only select descriptions with at least $500$ characters.

\begin{figure*}[h]
    \centering
    \includegraphics[width=\linewidth]{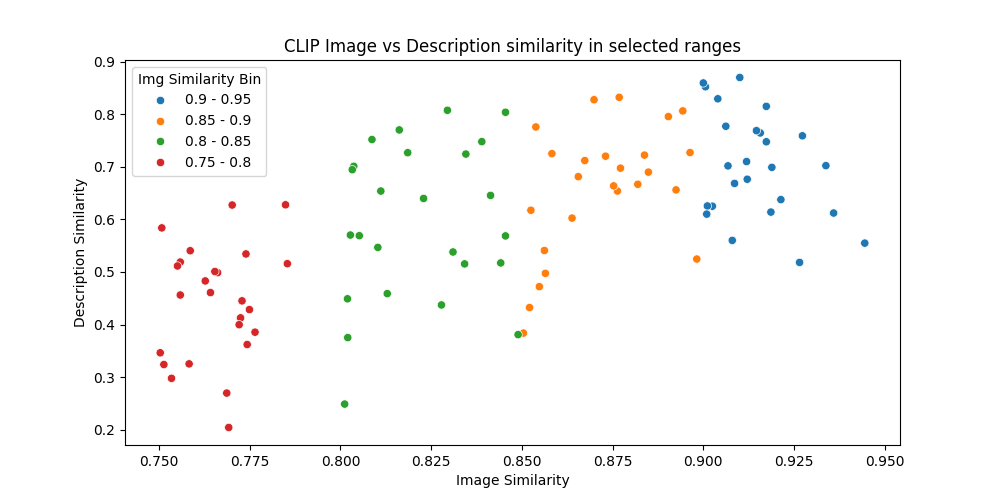}
    \caption{We plot the CLIP similarity between descriptions and images of the sampled example pairs. We find as expected that there is some positive correlation between the similarity of image pairs and the similarity of textual descriptions. However, we can also observe that some description similarity can be low even for images pairs which are predicted to be similar. This is because some of the descriptions are short and/or the annotators decided to focus on different aspects of the image.}
    \label{fig:clip-sim}
\end{figure*}

After sampling from our desired image CLIP similarity range, we plot our subset against the text CLIP similarity between each pairs of examples (shown in Figure~\ref{fig:clip-sim}). 

\begin{figure*}[h]\centering
\begin{tabular}{|p{0.45\textwidth}|p{0.45\textwidth}|}\hline\tiny{Outdoor medium shot view of the General W. K Wilson Jr. Bridge from a 3/4 view from behind the glass of a motor vehicle on the opposite road. There are droplets of water on the glass that are out of focus. There are dark gray rain clouds outside. A bridge arch over a portion of a highway road with suspension cords coming down from the left and right sides of the bridge where the beams cross horizontally on the arch. The archway resembles a silver ladder. Yellow reflective bumpers border the road on the lower right. A cement barrier rises between the arch and just before it.} & \tiny{An aerial view of a dark green and gray blue landscape with a river running through it. The image is low resolution and not in focus. The river is wide and runs from the bottom left corner to one third of the way up and out on the right edge of the frame. One tanker ship is traveling in the center of the river to the left and angled to the bottom left corner. A large sand bank bows out from the lower river bank as the river bends to the right. Below the lower river bank is a forested area with many thick trees. A tributary river feeds the main river from the right, and meanders down to the left. Above the far side of the river, a forest makes a large loop and fills the center of the frame. Farm land fills the top half of the frame beyond.} \\ \hline\end{tabular}
\includegraphics[width=0.4\textwidth]{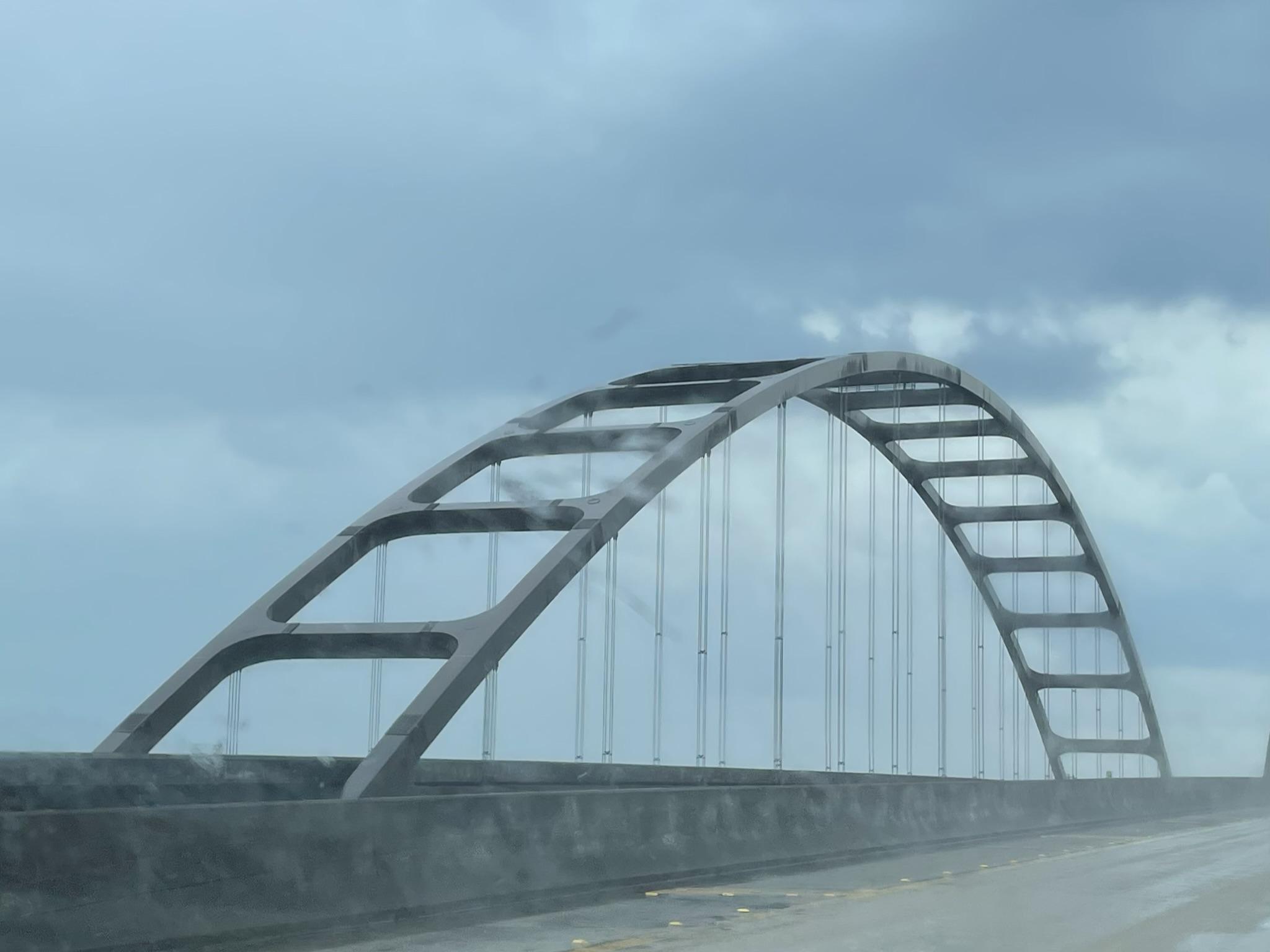}
\includegraphics[width=0.4\textwidth]{example-images/4_1.jpg}
\tiny{\begin{tabular}{|c|p{0.85\textwidth}|}
\hline
\hline
Bunny 1.1 & 1. Both images feature a bridge structure.

2. The bridge in both images has a curved arch design.

3. The color of the bridge is gray.

4. The sky in both images is blue.

5. The clouds in both images are white. \\ 
\hline
LLaVA 1.5 (Vicuna) & 1. Both images feature a bridge.

2. Both images show a body of water.

3. Both images depict a cloudy sky.

4. Both images have a similar color palette, with blues and grays dominating.

5. Both images show a landscape with natural elements.  \\ 
\hline
MiniCPM V2.5 & 1. Both images are taken during the daytime.

2. They both feature man-made structures.

3. The sky is visible in both images.

4. There is a sense of motion or travel suggested by the road in the left image and the aerial view in the right image.

5. Both images have a color palette dominated by shades of blue and gray. \\ 
\hline
InternVL2 & 1. Both images feature a bridge.

2. Both images have a cloudy sky.

3. Both images show a landscape with water bodies.

4. Both images have a sense of depth and perspective.

5. Both images have a similar color palette, with blues and grays dominating the scene. \\ 
\hline
GPT4o-Mini & Here are five similarities between the two images:

1. Curved Structures: Both images feature prominent curved shapes; the left image shows an arch bridge, while the right image depicts a river with a winding curve.

2. Natural Elements: Each image incorporates natural elements; the right image prominently features a river and land, while the left image is set against a cloudy sky.

3. Man-Made vs. Natural: The left image represents a man-made structure (the bridge), while the right image showcases a natural landscape, highlighting the contrast between human engineering and nature.

4. Perspective: Both images are \\ 
\hline
Phi 3.5 Vision & 1. Both images feature man-made structures.

2. The first image shows a bridge, while the second image shows a river with a bridge.

3. Both images have a blue color palette.

4. The first image has a cloudy sky, while the second image has a clear sky.

5. Both images are taken from a ground perspective. \\ 
\hline
LLaVA 1.6 (Llama) & 1. Both images show a bridge structure.

2. The bridge in both images appears to be a suspension bridge.

3. The bridge is located over a body of water.

4. The sky in both images is partly cloudy.

5. The bridge is the main focus of both images, with the water and sky serving as the background. \\ 
\hline
\end{tabular}}
\vspace{10mm}
\caption{Generated statements for each model when given \textcolor[HTML]{2f6eba}{image} inputs}
\label{fig:example_4}
\end{figure*}

\begin{figure*}[h]\centering
\begin{tabular}{|p{0.45\textwidth}|p{0.45\textwidth}|}\hline\tiny{An eye-level view of a tree trunk that has been ripped out of the ground laying on its side. The tree trunk is facing away from view, only the bottom and very top of the tree trunk is visible. The bottom of the tree trunk is hollow, there is a hole visible through the bottom that allows you to see a small sliver of the ground in the distance. The ground is sloped toward the bottom left corner of the image, it is a dirt surface that is covered mostly with gray discolored leaves and brown leaves scattered throughout the image. There are thin tree trunks and trees in the background behind the tree trunk.} & \tiny{An overhead view of a group of nine California pipevine swallowtail butterflies sitting on a dirt surface. The butterflies are all facing different directions. The front of their wings are dark blue and fade into a lighter shade of blue as they go back. There are white dots lining the edge of each butterfly's wings. There are two large gray rocks visible in the bottom right and bottom left corner of the image. A large concentration of dry leaves and sticks are covering the dirt surface at the top half of the image. There are sticks and dry leaves scattered more sparingly in the middle of the image where the butterflies are standing.} \\ \hline\end{tabular}
\includegraphics[width=0.4\textwidth]{example-images/15_0.jpg}
\includegraphics[width=0.4\textwidth]{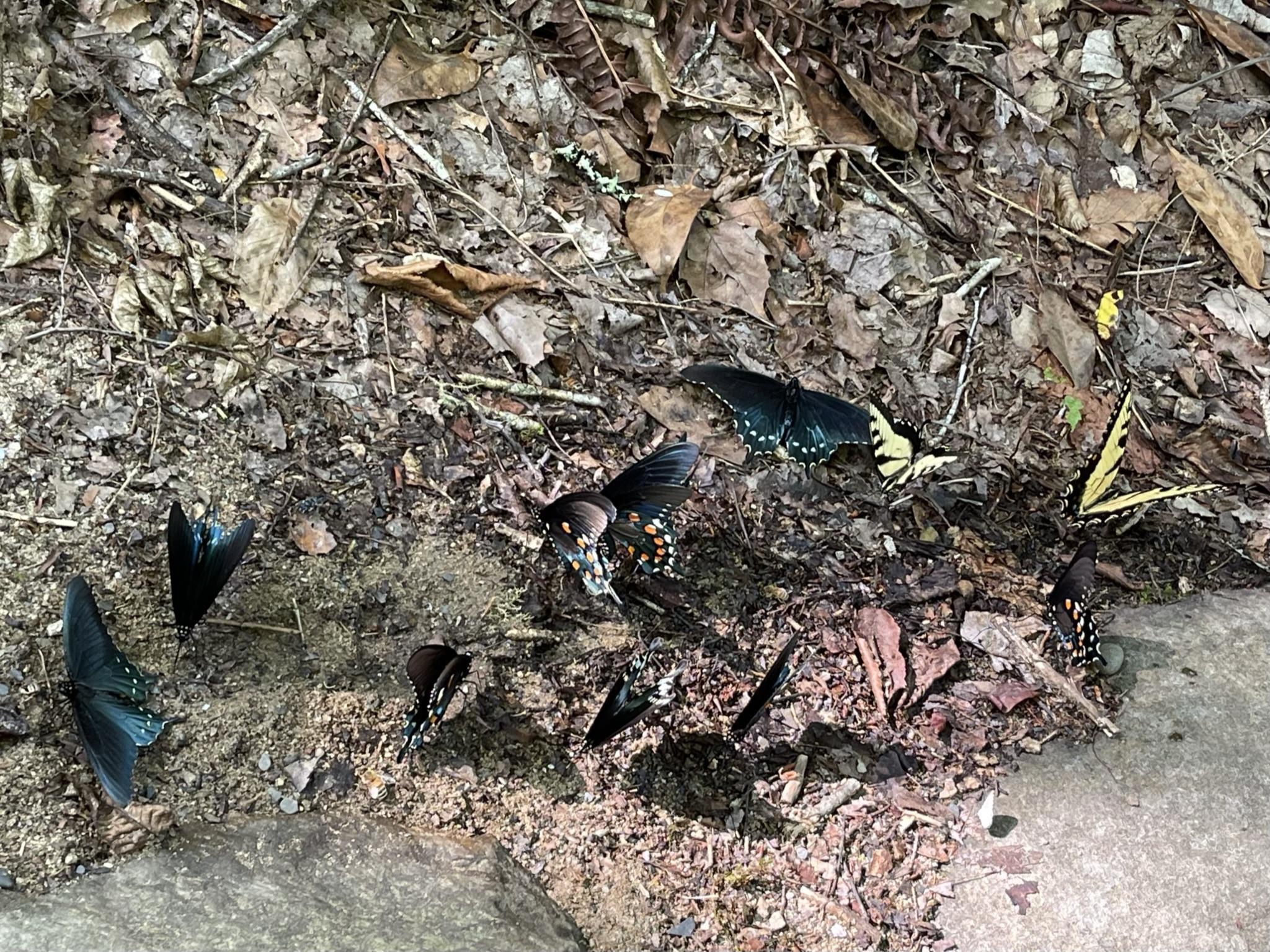}
\tiny{\begin{tabular}{|c|p{0.85\textwidth}|}
\hline
\hline
Bunny 1.1 & 1. Both images feature a fallen tree stump.

2. The tree stump is surrounded by a mix of leaves and twigs.

3. There are butterflies present in both images.

4. The butterflies are black with yellow spots.

5. The background of both images is a forest with green foliage. \\ 
\hline
LLaVA 1.5 (Vicuna) & 1. Both images show a natural setting with a forest floor.

2. There is a fallen tree trunk in both images.

3. The fallen tree trunks are surrounded by leaves and debris.

4. The ground appears to be damp and muddy.

5. The overall color palette is earthy and muted.  \\ 
\hline
MiniCPM V2.5 & 1. The presence of butterflies is a common element in both images.

2. Both images depict a natural outdoor setting with trees and foliage.

3. The ground in both images is covered with leaves and twigs, indicating a forest floor environment.

4. The lighting in both images suggests they were taken during the daytime.

5. The overall composition of both images focuses on the interaction between the butterflies and their natural habitat. \\ 
\hline
InternVL2 & 1. The presence of a large tree stump in the center of each image.

2. The ground is covered with fallen leaves and branches.

3. The images show a forested area with trees and foliage.

4. The lighting in both images appears to be natural, suggesting they were taken during the day.

5. The images have a similar color palette, with earthy tones dominating the scene. \\ 
\hline
GPT4o-Mini & Here are five similarities between the two images:

1. Natural Setting: Both images are set in a natural environment, showcasing elements of the outdoors.

2. Organic Materials: Each image features organic materials; the left image has a decaying tree stump, while the right image includes butterflies interacting with the ground.

3. Vegetation: Both images contain vegetation, with leaves and plants visible in the surroundings.

4. Earthy Tones: The color palette in both images includes earthy tones, such as browns, greens, and muted colors.

5. Wildlife Presence: Each image depicts wildlife \\ 
\hline
Phi 3.5 Vision & Both images depict natural outdoor scenes.

2. Each image contains organic debris, such as leaves and branches.

3. The color palette is dominated by earthy tones like browns and greens.

4. Both images have a ground-level perspective, looking upwards towards the sky.

5. There is a sense of decay and natural processes at work in both images. = 0.00000000000000000000000000 \\ 
\hline
LLaVA 1.6 (Llama) & 1. Both images show a natural outdoor setting with a focus on a fallen tree or stump.

2. The ground appears to be covered with leaves and possibly moss, indicating a forested or wooded area.

3. There are insects or small creatures on the ground near the tree stump in both images.

4. The lighting suggests it might be daytime with natural sunlight.

5. The overall composition of the images is similar, with the tree stump as the central point of interest. \\ 
\hline
\end{tabular}}
\vspace{10mm}
\caption{Generated statements for each model when given \textcolor[HTML]{2f6eba}{image} inputs}
\label{fig:example_15}
\end{figure*}

\begin{figure*}[h]\centering
\begin{tabular}{|p{0.45\textwidth}|p{0.45\textwidth}|}\hline\tiny{A group of four Tiger sharks under water in an aquarium, the sharks appear to be near some man made stones and a small school of fish to the bottom left corner. The sharks have grey skin with white pale underbellies, majority of the sharks are facing towards the left. In the center is a rock where there is a shark in front of the rock and one behind it, the shark to the front of the rock is facing to the right. The water is dark blue with a light shining to the left side of the photo and some reflections on the surface below.} & \tiny{A low-angle view of three sharks swimming in an aquarium among a large number of small gray fish scattered throughout the image. One of the sharks is on the top left side of the image swimming toward the top right corner of the image. There is another shark further away on the right side of the image facing the left side of the image. The front half of a shark is visible extending from the bottom right side of the image facing the left side of the image. There is a light blue hue throughout the image and the water in the distance fades into blue. Light from above the surface of the water is visible at the top of the image shining through the ripples on the surface of the water. The light is shining on the top shark and on the fish at the top of the image.} \\ \hline\end{tabular}
\includegraphics[width=0.4\textwidth]{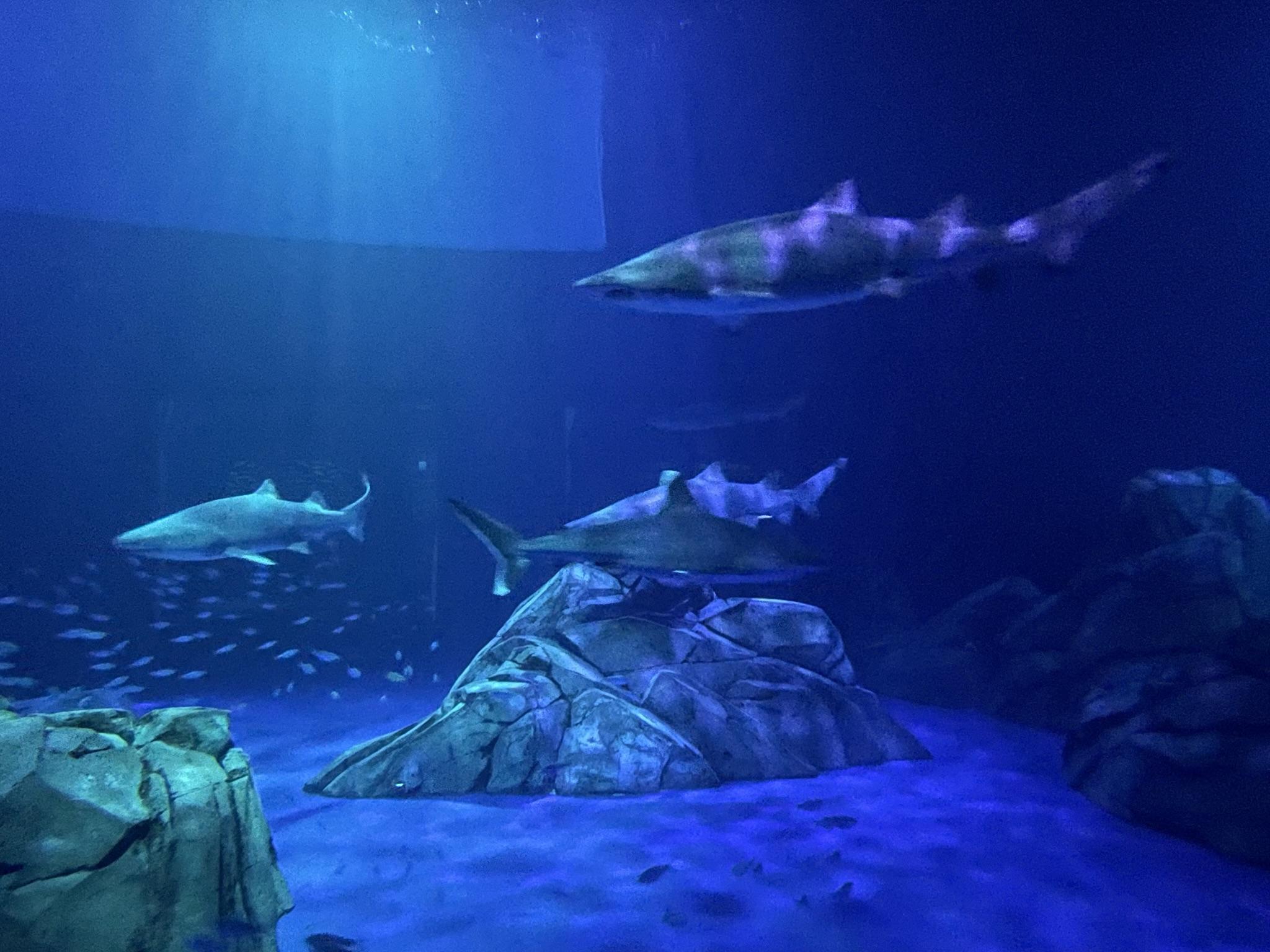}
\includegraphics[width=0.4\textwidth]{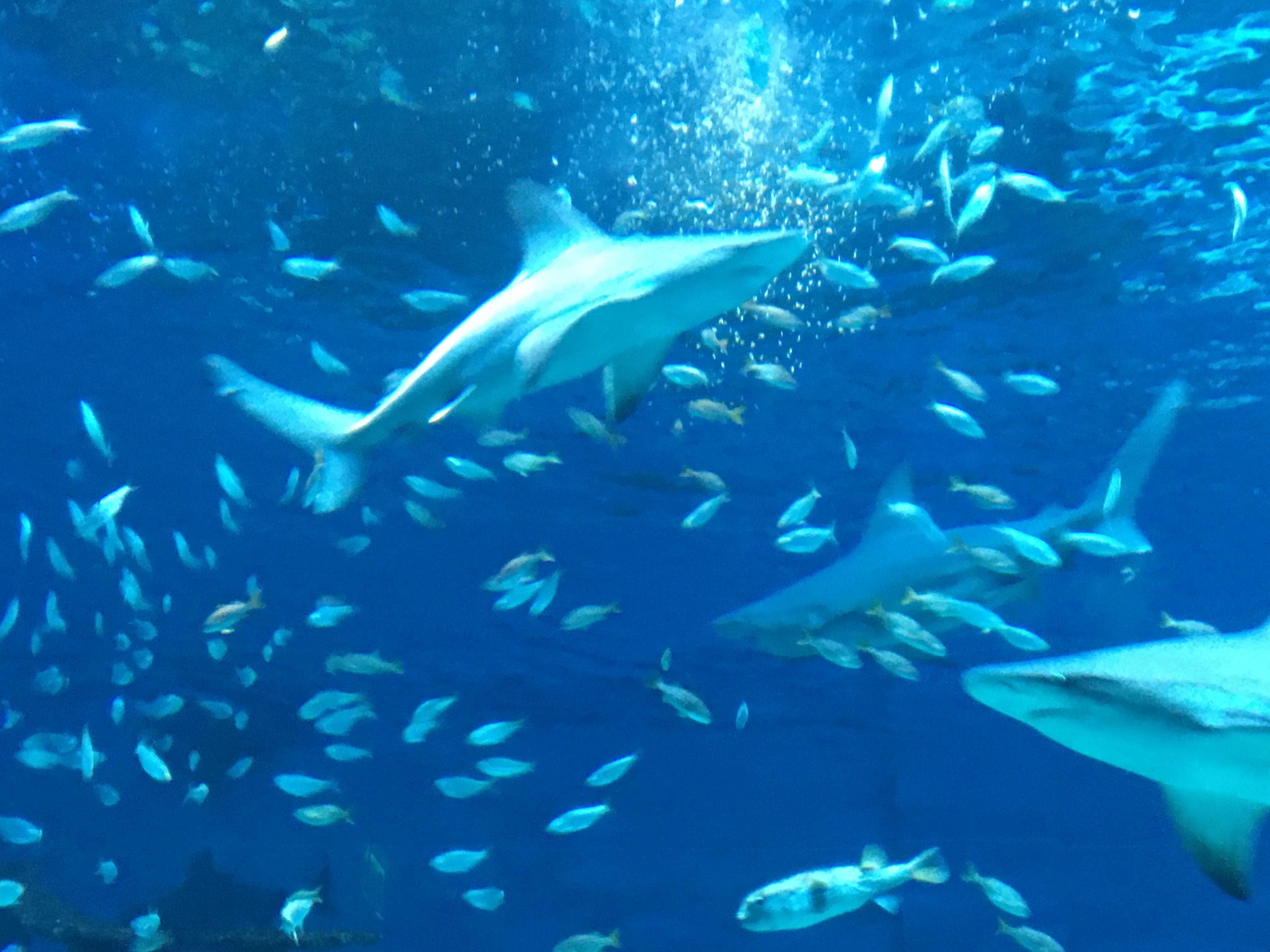}
\tiny{\begin{tabular}{|c|p{0.85\textwidth}|}
\hline
\hline
Bunny 1.1 & 1. Both scenes take place in an aquarium.

2. Both scenes feature sharks as the main subjects.

3. Both scenes have a school of fish present.

4. Both scenes have a light source that is shining on the water surface.

5. Both scenes have a dark blue hue to the water. \\ 
\hline
LLaVA 1.5 (Vicuna) & 1. Both scenes feature sharks in an aquarium environment.

2. Both scenes depict sharks swimming near rocks and fish.

3. The sharks in both scenes have grey skin with white underbellies.

4. The majority of the sharks in both scenes are facing towards the left.

5. There is a light shining in the water in both scenes, creating reflections on the surface.  \\ 
\hline
MiniCPM V2.5 & 1. Both scenes take place in an aquarium.

2. Both scenes feature sharks swimming in the water.

3. Both scenes have a school of fish present.

4. Both scenes have a light source shining on the water surface.

5. Both scenes have reflections on the water surface. \\ 
\hline
InternVL2 & The sharks are in an aquarium.

2. The sharks are swimming.

3. The sharks are surrounded by fish.

4. The water is blue.

5. There is light shining on the sharks. \\ 
\hline
GPT4o-Mini & Here are five similarities between Scene 1 and Scene 2:

1. Aquarium Setting: Both scenes take place in an aquarium environment, showcasing sharks swimming in water.

2. Presence of Sharks: Each scene features multiple sharks; Scene 1 has four Tiger sharks, while Scene 2 has three sharks.

3. Interaction with Fish: In both scenes, the sharks are surrounded by small fish. Scene 1 has a small school of fish in the bottom left corner, while Scene 2 has a large number of small gray fish scattered throughout the image.

4. Lighting Effects: Both scenes exhibit \\ 
\hline
Phi 3.5 Vision &1. Both scenes feature sharks in an aquarium setting.

2. Both scenes feature a group of sharks.

3. Both scenes feature small fish in the background.

4. Both scenes feature a light source shining on the sharks.

5. Both scenes feature a dark blue color palette.

 \\ 
\hline
LLaVA 1.6 (Llama) & 1. Both scenes take place in an aquarium.

2. Both scenes feature sharks.

3. Both scenes feature fish.

4. Both scenes have a water setting.

5. Both scenes have a light source shining on the water. \\ 
\hline
\end{tabular}}
\vspace{10mm}
\caption{Generated statements for each model when given \textcolor[HTML]{f09737}{text} inputs}
\label{fig:example_92}
\end{figure*}

\begin{figure*}[h]\centering
\begin{tabular}{|p{0.45\textwidth}|p{0.45\textwidth}|}\hline\tiny{A close up view of a large glass sphere light fixture and a yellow light, hanging freely from a silver metal pole. To the right of the light pole, is a gold metal pole with a "M" next to it on the left. The light pole is creating a shadow behind it. There is a green and yellow wall behind the light fixtures, the green wall is on the left, and the yellow wall is on the right. There is a black wire and push bottom hanging behind the light fixture. To the left of the sphere light fixture, is a screen glass door leading to outside. Plants can be seen on the bottom left and right corners.} & \tiny{A low-angle shot of a creative light blue pendant lamp on a gray concrete ceiling. In the center is the lamp made with light blue rattan material, creating a woven spherical shape with a view of a white cylinder case with a bright white light in the center. The lamp is connected to a silver metal rod and a silver metal cylinder adjacent to the ceiling above. Behind the lamp is a gray concrete ceiling made with cement. The ceiling is vaulted diagonally from left to right. On the bottom of the frame is a gray tubing system of pipes connected to the ceiling.} \\ \hline\end{tabular}
\includegraphics[width=0.4\textwidth]{example-images/66_0.jpg}
\includegraphics[width=0.4\textwidth]{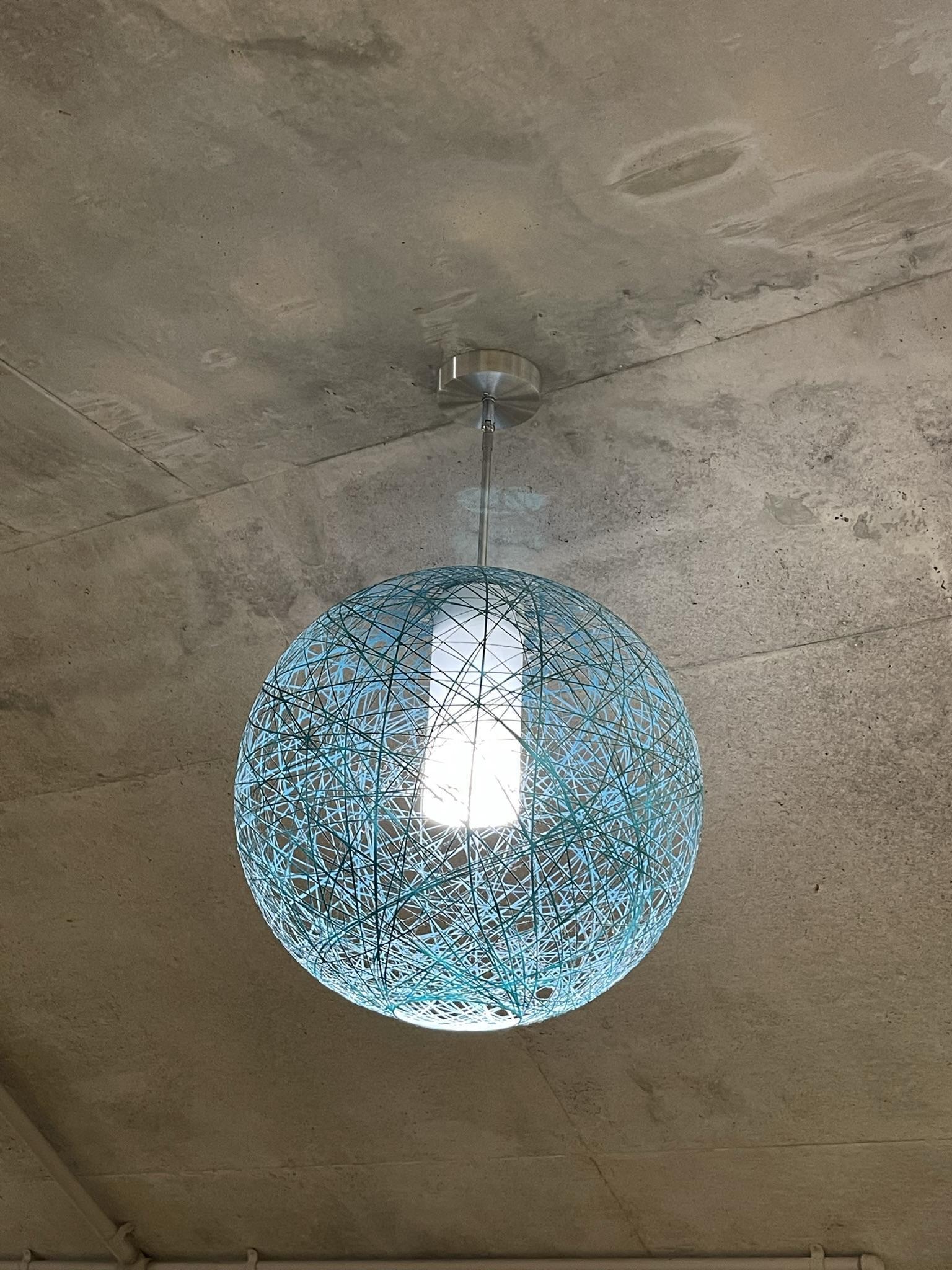}
\tiny{\begin{tabular}{|c|p{0.85\textwidth}|}
\hline
\hline
Bunny 1.1 & 1. Both scenes feature a light fixture hanging from a pole.

2. Both scenes have a wall in the background.

3. Both scenes have a shadow being cast by the light fixture.

4. Both scenes have a wire hanging behind the light fixture.

5. Both scenes have a door leading to outside. \\ 
\hline
LLaVA 1.5 (Vicuna) & 1. Both scenes feature a light fixture hanging from a pole.

2. Both scenes have a yellow light.

3. Both scenes have a silver metal pole.

4. Both scenes have a green and yellow wall.

5. Both scenes have a black wire and push bottom hanging behind the light fixture.  \\ 
\hline
MiniCPM V2.5 & 1. Both scenes have a light fixture as the main focus.

2. Both scenes have a metal pole or rod supporting the light fixture.

3. Both scenes have a shadow created by the light fixture.

4. Both scenes have a wall in the background.

5. Both scenes have a door or opening leading to the outside. \\ 
\hline
InternVL2 & 1. Both scenes have a light fixture.

2. Both scenes have a metal pole.

3. Both scenes have a shadow.

4. Both scenes have a green wall.

5. Both scenes have a black wire. \\ 
\hline
GPT4o-Mini & Here are the similarities between Scene 1 and Scene 2:

1. Light Fixtures: Both scenes feature prominent light fixtures that are central to the composition—Scene 1 has a large glass sphere light fixture, while Scene 2 showcases a light blue pendant lamp.

2. Metal Components: Each scene includes silver metal elements—Scene 1 has a silver metal pole, and Scene 2 has a silver metal rod and cylinder.

3. Spherical Shape: Both light fixtures have a spherical design; Scene 1's glass sphere and Scene 2's woven spherical lamp.

4. Background Elements: \\ 
\hline
Phi 3.5 Vision & 1. Both scenes feature a light source hanging from a pole or rod.

2. Both scenes have a metallic element in the form of a pole or rod.

3. Both scenes have a wall in the background.

4. Both scenes have a spherical element in the form of a light fixture or pendant lamp.

5. Both scenes have a green element in the form of a plant or foliage.

 \\ 
\hline
LLaVA 1.6 (Llama) & 1. Both scenes feature a light fixture hanging from a pole.

2. Both scenes have a wall in the background.

3. Both scenes have a shadow behind the light fixture.

4. Both scenes have a wire and push button hanging behind the light fixture.

5. Both scenes have plants in the bottom left and right corners. \\ 
\hline
\end{tabular}}
\vspace{10mm}
\caption{Generated statements for each model when given \textcolor[HTML]{f09737}{text} inputs}
\label{fig:example_66}
\end{figure*}

\begin{figure*}[h]\centering
\begin{tabular}{|p{0.45\textwidth}|p{0.45\textwidth}|}\hline\tiny{A Lamborghini showroom is seen from outside the window, which shows reflections of lights outside. A bright blue Lamborghini is seen in the front, with a silver convertible to the right and a white SUV in the background. The blue car in the front is seen from the front passenger's side, with a white note on the windshield and a large white tag hanging from the rearview mirror. Several small lights are reflected on this vehicle, while the silver next to it shows one large reflection over the front passenger fender. Behind the white SUV is a doorway leading to a lit office. Behind the other two vehicles is a large black window on the left, indicating nighttime, two small racks of t-shirts, and a set of stairs leading up to the right. The flooring in the showroom is made of large concrete slabs.} & \tiny{A front view of a red Lamborghini Aventador parked in the middle of a grey concrete show room floor. The Lamborghini has a black grill and the Lamborghini logo below the hood of the car. Reflections of light are on the hood of the Aventador and left headlight. A shadow of the Lamborghini encircles the front end of the car on the ground. A neon Lamborghini sign hangs on the wall in the background in the right corner behind a white Lamborghini car. A white couch and clothes hanging on a rack are behind the red Lamborghini on the left. A black staircase is behind the red Lamborghini in the background.} \\ \hline\end{tabular}
\includegraphics[width=0.4\textwidth]{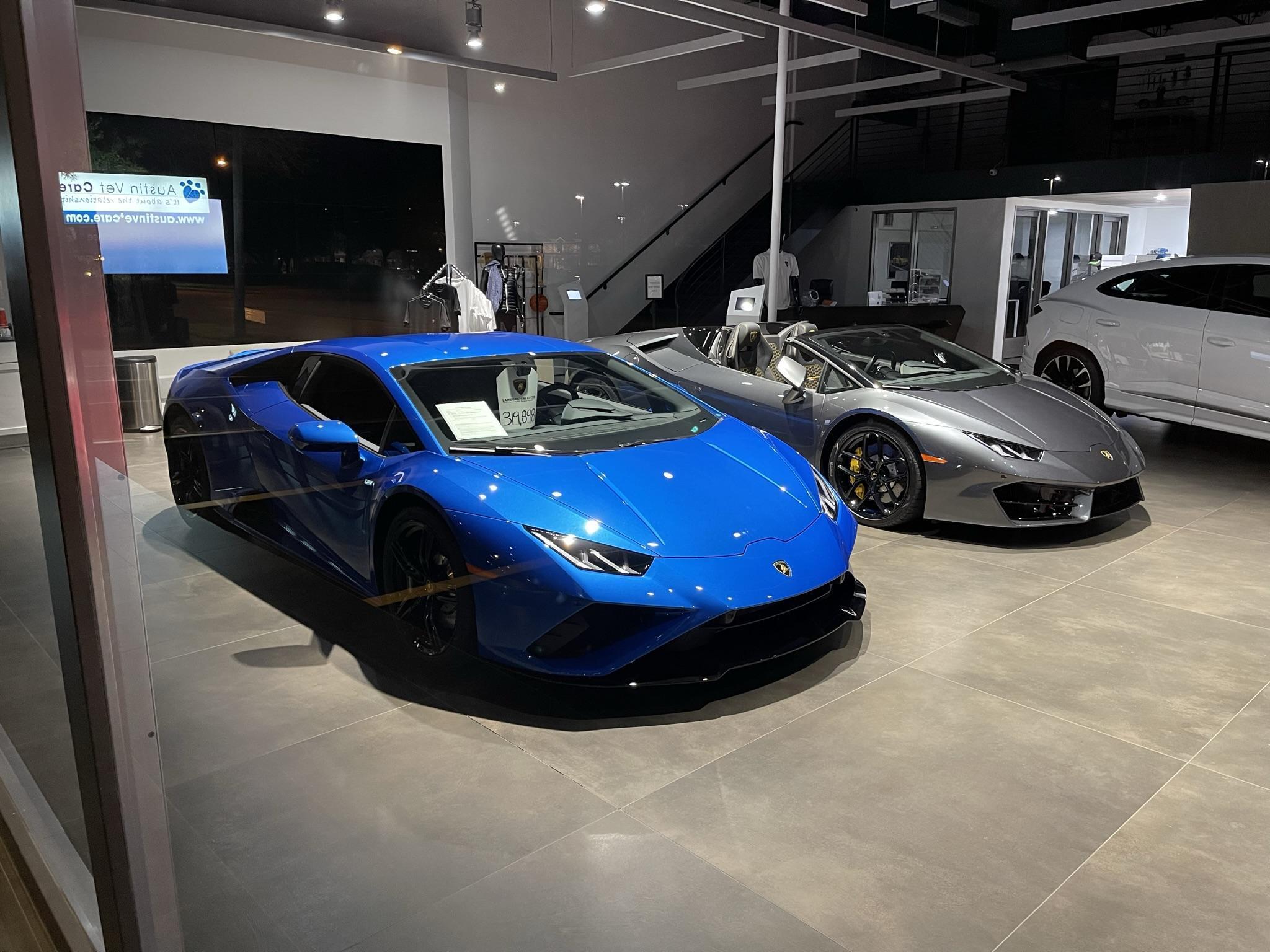}
\includegraphics[width=0.4\textwidth]{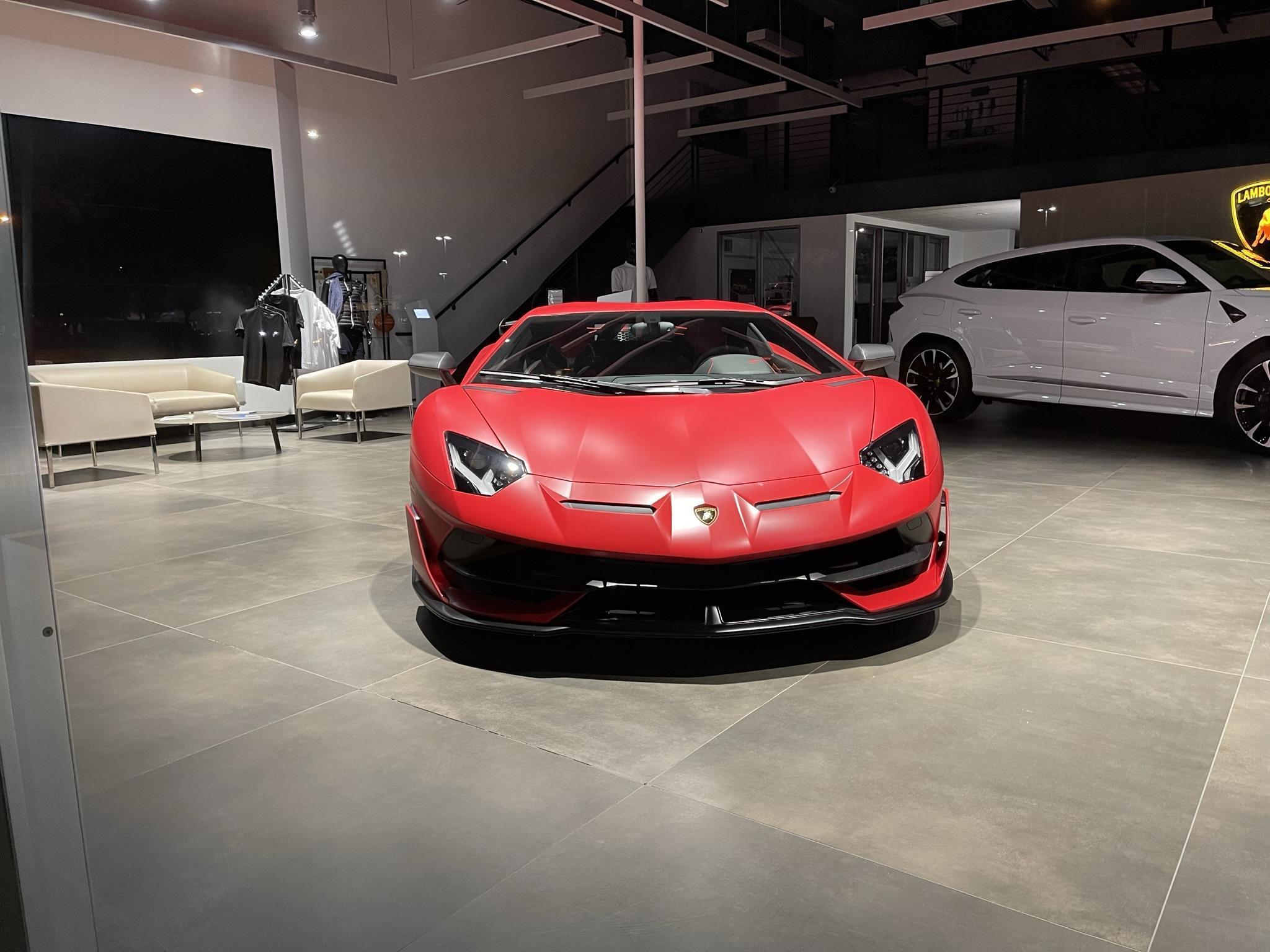}
\tiny{\begin{tabular}{|c|p{0.85\textwidth}|}
\hline
\hline
Bunny 1.1 & 1. Both scenes feature a Lamborghini showroom.

2. The showrooms have a concrete floor.

3. There are reflections of lights on the cars.

4. There is a Lamborghini logo visible on the cars.

5. There is a staircase in the background of both scenes. \\ 
\hline
LLaVA 1.5 (Vicuna) & 1. Both scenes feature a Lamborghini showroom.

2. Both scenes have a variety of Lamborghini cars on display.

3. Both scenes have a mix of luxury and sports cars.

4. Both scenes have a modern and sleek interior design.

5. Both scenes have a focus on the cars, with the cars being the main subject.  \\ 
\hline
MiniCPM V2.5 & 1. Both scenes feature Lamborghini vehicles prominently displayed in showrooms.

2. The showrooms have a modern and sleek design with concrete flooring.

3. Reflections of lights are visible on the vehicles in both scenes.

4. There is a white Lamborghini in the background of both scenes.

5. Both scenes include a black staircase in the background. \\ 
\hline
InternVL2 & 1. Both scenes feature a Lamborghini showroom.

2. Both scenes have a concrete floor.

3. Both scenes have a Lamborghini car prominently displayed.

4. Both scenes have a staircase in the background.

5. Both scenes have a large window indicating nighttime. \\ 
\hline
GPT4o-Mini & Here are five similarities between the two scenes:

1. Showroom Environment: Both scenes take place in a Lamborghini showroom with a modern design and concrete flooring.

2. Car Models: Each scene features a Lamborghini as the focal point, showcasing different models (a blue Lamborghini in Scene 1 and a red Lamborghini in Scene 2).

3. Lighting and Reflections: Both images display reflections of lights on the cars, enhancing the visual appeal and highlighting the vehicles.

4. Additional Vehicles: Each scene includes multiple vehicles; Scene 1 has a silver convertible and a white SUV, while Scene 2 features \\ 
\hline
Phi 3.5 Vision & 1. Both scenes feature a Lamborghini car as the main subject.

2. Both images show a showroom setting with a clean, modern interior.

3. Each image has a car parked in the foreground with another car in the background.

4. Both images have a reflective surface on the car, indicating a polished finish.

5. The flooring in both images is made of large, flat concrete slabs.  \\ 
\hline
LLaVA 1.6 (Llama) & 1. Both scenes feature a Lamborghini showroom.

2. Both scenes have a Lamborghini car as the central focus.

3. Both scenes have a showroom floor with a concrete finish.

4. Both scenes have a car with a large grille and the Lamborghini logo.

5. Both scenes have reflections of light on the cars.

6. Both scenes have a neon sign with the Lamborghini logo.

7. Both scenes have a staircase in the background.

8. Both scenes have a white couch in the background.

9. Both scenes have a car with a large windshield.

10. Both scenes have a car with a black headlight. \\ 
\hline
\end{tabular}}
\vspace{10mm}
\caption{Generated statements for each model when given \textcolor[HTML]{4ead5b}{both} inputs}
\label{fig:example_68}
\end{figure*}

\begin{figure*}[h]\centering
\begin{tabular}{|p{0.45\textwidth}|p{0.45\textwidth}|}\hline\tiny{Paper lanterns are seen hanging from a tent-type ceiling with a metal frame. Four of the lanterns are globe-shaped, while the fifth one that hangs behind the bottom left globe is a flattened shape. Both of the globe lanterns on the right are lit up. All but one of the globes have a symmetrical frame. The top right globe has a swirled frame and has two tears on the bottom of the paper. The right bottom lantern has a tear on the side, and the bottom left lantern has a tear on the bottom. The ceiling is black cloth with large metal frame beams. A small blue sign hangs from the metal beam on the left and can only partially be seen in the bottom corner. The sign is a circular shape and reads "MENUS" in white.} & \tiny{A very large light fixture is seen from a low angle, attached to wood beams on a wood plank ceiling. It has a large black circle base with numerous black cords of different lengths attached. The black cords hold iridescent glass globes with open bottoms. Each globe has a warm-toned round light bulb lit up inside. The ceiling is made of dark wood with small flush-mounted lights. Behind the large light fixture is another large light fixture that is slightly different, with a hanging hollow circular base that has round glass globes attaching to the base with warm light bulbs lit up. Two white brick pillars are seen in the distance in the bottom right of the image.} \\ \hline\end{tabular}
\includegraphics[width=0.4\textwidth]{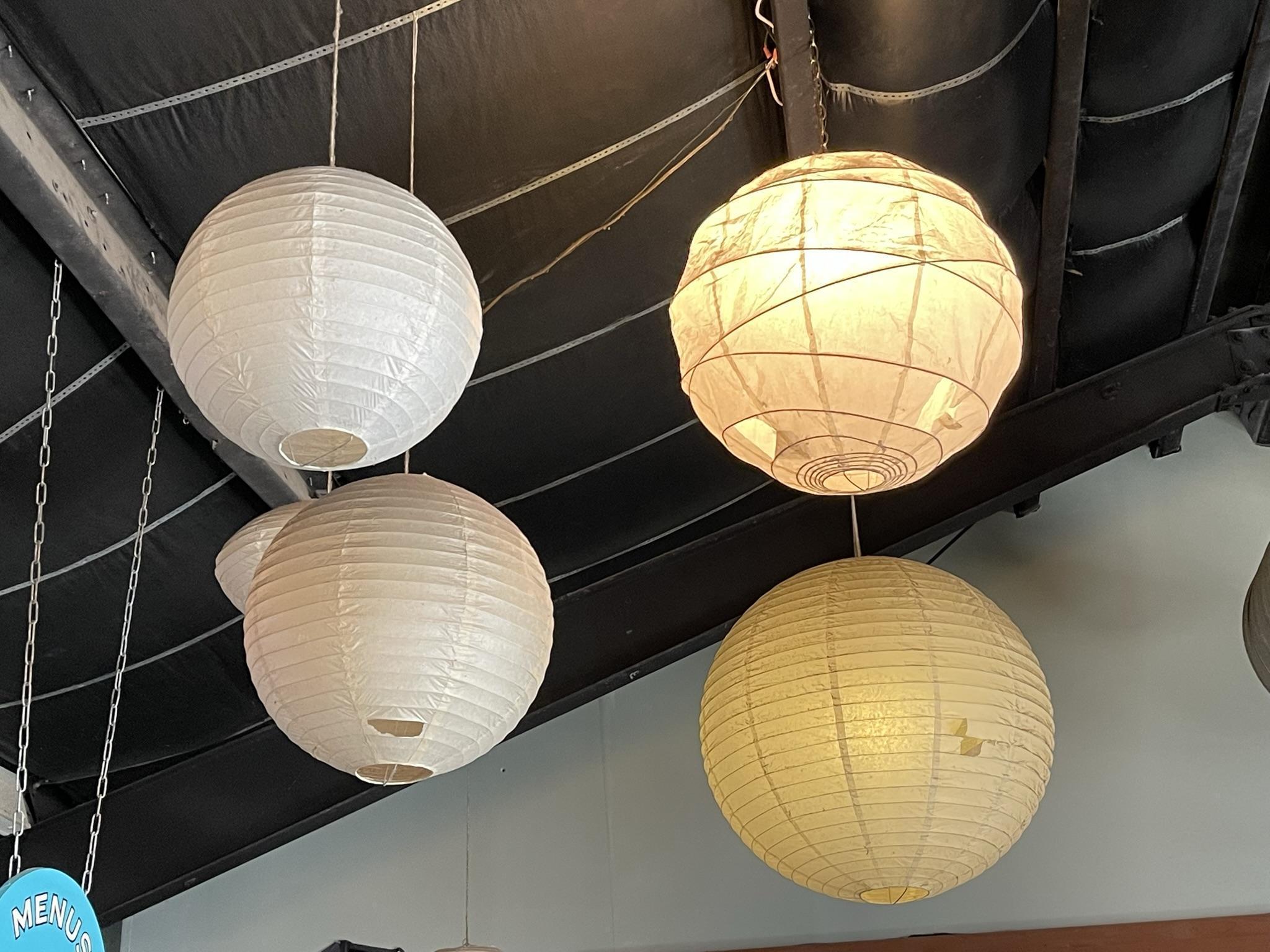}
\includegraphics[width=0.4\textwidth]{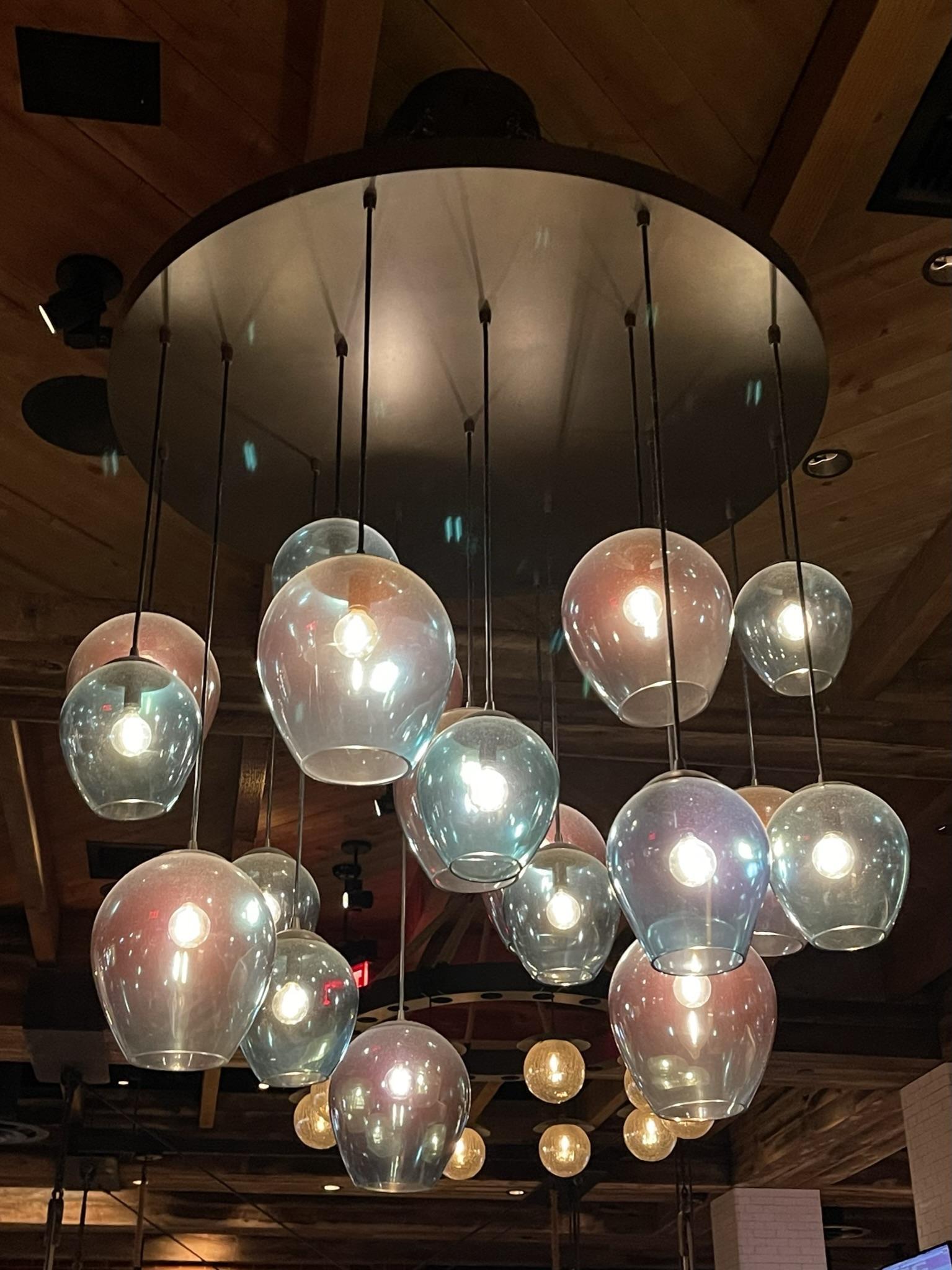}
\tiny{\begin{tabular}{|c|p{0.85\textwidth}|}
\hline
\hline
Bunny 1.1 & 1. Both scenes have a ceiling with a unique design.

2. Both scenes have light fixtures that are the main focus.

3. Both scenes have multiple light sources.

4. Both scenes have a warm-toned light bulb.

5. Both scenes have a circular shape in the light fixtures. \\ 
\hline
LLaVA 1.5 (Vicuna) & 1. Both scenes feature hanging light fixtures.

2. Both scenes have a combination of globe-shaped and flattened light fixtures.

3. Both scenes have light bulbs lit up inside the fixtures.

4. Both scenes have a symmetrical frame design for the globe-shaped light fixtures.

5. Both scenes have a black ceiling or background.  \\ 
\hline
MiniCPM V2.5 & 1. Both scenes feature hanging light fixtures with multiple globes.

2. The globes in both scenes are illuminated.

3. The ceiling in both scenes is dark, with the first being cloth and the second being wood.

4. Both scenes have a circular base for the light fixtures.

5. The globes in both scenes have an open bottom design. \\ 
\hline
InternVL2 & 1. Both scenes feature hanging light fixtures.

2. Both scenes have a ceiling with a dark color.

3. Both scenes have light bulbs lit inside the fixtures.

4. Both scenes have a variety of shapes for the hanging elements.

5. Both scenes have a sense of a cozy, warm atmosphere. \\ 
\hline
GPT4o-Mini & Here are five similarities between the two scenes:

1. Hanging Light Fixtures: Both scenes feature light fixtures that are suspended from the ceiling, creating a focal point in the space.

2. Globe Shapes: Each scene includes globe-shaped elements; Scene 1 has globe-shaped paper lanterns, while Scene 2 has iridescent glass globes.

3. Illumination: Both scenes showcase light sources that are lit up, contributing to the ambiance of the environment.

4. Ceiling Design: The ceilings in both scenes are designed with a dark color scheme—black cloth in Scene 1 \\ 
\hline
Phi 3.5 Vision & 1. Both scenes feature hanging light fixtures.

2. The light fixtures in both scenes are made of glass globes.

3. Both light fixtures have a warm-toned light bulb inside each globe.

4. The light fixtures are suspended from a ceiling structure.

5. Both light fixtures have a modern and artistic design.  \\ 
\hline
LLaVA 1.6 (Llama) & 1. Both scenes feature light fixtures with multiple glass globes.

2. The glass globes in both scenes have light bulbs inside.

3. Both scenes have a mix of round and non-round glass globes.

4. Both scenes have a ceiling with a combination of wood and metal elements.

5. Both scenes have a warm lighting effect from the bulbs inside the glass globes. \\ 
\hline
\end{tabular}}
\vspace{10mm}
\caption{Generated statements for each model when given \textcolor[HTML]{4ead5b}{both} inputs}
\label{fig:example_63}
\end{figure*}

\end{document}